\documentclass[11pt]{article}
\usepackage[papersize={8.5in,11in},margin=1in]{geometry}

\usepackage[utf8]{inputenc}
\usepackage[T1]{fontenc}
\usepackage{xspace}
\usepackage{cprotect}

\usepackage{algorithm}
\usepackage{algpseudocode}
\usepackage{algorithmicx}
\usepackage{amsthm}
\usepackage{mathrsfs}
\usepackage{subcaption}

\usepackage{natbib}
\usepackage{graphicx}
\usepackage{url}
\usepackage{amssymb}
\usepackage{bm
}\usepackage{amsmath}
\usepackage[hidelinks]{hyperref}
\hypersetup{
colorlinks=true,
citecolor=blue,
}
\usepackage{dirtytalk}
\usepackage{enumitem}

\usepackage[hyperpageref]{backref}
\newcommand{\comfyfill}[1]{%
  \unskip\hspace*{0.1em plus 1fill}
  \nolinebreak[3]%
  \hspace*{\fill}\mbox{#1}
  \parfillskip0pt\par
}
\newcommand\foo[1]{{\comfyfill{\mbox{#1}}}}
\renewcommand*{\backref}[1]{}
\renewcommand*{\backrefalt}[4]{%
\ifcase #1 %
\or
\foo{(page #2)}%
\else
\foo{(pages #2)}%
\fi
}

\usepackage{cleveref}

\setitemize{noitemsep,topsep=0pt,parsep=0pt,partopsep=0pt}
\setenumerate{noitemsep,topsep=0pt,parsep=0pt,partopsep=0pt}

\usepackage[backgroundcolor=White,textwidth=3cm]{todonotes}

\def\advertorch{{\fontfamily{fco}{\fontseries{l}\selectfont advertorch}}\xspace}

\title{\advertorch v0.1: An Adversarial Robustness \\Toolbox based on PyTorch}
\author{Gavin Weiguang Ding, Luyu Wang, Xiaomeng Jin\\
\\
Borealis AI
}
\date{}

\begin{document}
\maketitle

\begin{abstract}
\advertorch is a toolbox for adversarial robustness research. It contains various implementations for attacks, defenses and robust training methods.
\advertorch is built on PyTorch \citep{paszke2017automatic}, and leverages the advantages of the dynamic computational graph to provide concise and efficient reference implementations.
The code is licensed under the LGPL license and is open sourced at \url{https://github.com/BorealisAI/advertorch}.

\end{abstract}

\section{Introduction}

Machine learning models are vulnerable to ``adversarial'' perturbations \citep{szegedy2013intriguing,biggio2013data}. They are adversarial in the sense that, after these artificially constructed perturbations are added to on the inputs of the model, human observers do not change their perception, but the predictions of a model could be manipulated. Efforts of adversarial robustness research can be roughly divided into the following categories: generating strong and efficient attacks \citep{goodfellow2014explaining,carlini2017towards,brendel2017decision,xiao2018spatially,dong2018boosting}; detecting adversarial examples \citep{metzen2017detecting,ma2018characterizing,feinman2017detecting}; defending already trained models \citep{xu2017feature0,guo2017countering}; training robust models \citep{kurakin2016adversarial,madry2017towards,cisse2017parseval,ding2018max,wong2018provable,mirman2018differentiable}; robustness evaluation methodologies \citep{weng2018evaluating,katz2017reluplex,athalye2018obfuscated}; and understanding the vulnerability phenomena \citep{fawzi2017classification,shafahi2018adversarial,ding2018on}.

\advertorch aims to provide researchers the tools for conducting research in all the above mentioned directions. The current version of \advertorch include three of these aspects: attacks, defenses and robust training.
Compared to existing adversarial robustness related toolboxes \citep{papernot2016transferability,rauber2017foolbox}, \advertorch aims for
\begin{enumerate}
\item \emph{simple} and \emph{consistent} APIs for attacks and defenses;
\item \emph{concise} reference implementations, utilizing the dynamic computational graphs in PyTorch; and
\item \emph{fast} executions with GPU-powered PyTorch implementations, which are important for ``attack-in-the-loop'' algorithms, e.g. adversarial training.
\end{enumerate}

In this technical report, we give an overview of the design considerations and implementations of attacks in \Cref{sec:attacks}, defenses and robust training in \Cref{sec:defense}, and the versioning system in \Cref{sec:version}.

\section{Attacks}
\label{sec:attacks}

\advertorch implements various types of attacks, where
\cprotect{\href{https://github.com/BorealisAI/advertorch/blob/master/advertorch/attacks/__init__.py}}{\verb|attacks/__init__.py|}
maintains a complete list of them. We outline some of our design considerations and choices when implementing these attacks. Specifically, we describe gradient-based attacks
in \Cref{sec:grad_attack}, other attacks in \Cref{sec:other_attack}, and the wrapper for Backward Pass Differentiable Approximation (BPDA) \citep{athalye2018obfuscated} in \Cref{sec:bpda}.

\subsection{Gradient-Based Attacks}
\label{sec:grad_attack}

\def\perturb{\texttt{perturb}\xspace}
\def\predict{\texttt{predict}\xspace}
\def\loss{\texttt{loss\_fn}\xspace}
\def\x{\texttt{x}\xspace}
\def\y{\texttt{y}\xspace}

\advertorch currently implements the following gradient-based attacks:
\begin{itemize}
\item \texttt{GradientAttack}, \texttt{GradientSignAttack} \citep{goodfellow2014explaining},
\item \texttt{L2BasicIterativeAttack}, \texttt{LinfBasicIterativeAttack} \citep{kurakin2016adversarial},
\item \texttt{LinfPGDAttack}, \texttt{L2PGDAttack} \citep{madry2017towards},
\item \texttt{CarliniWagnerL2Attack} \citep{carlini2017towards},
\item \texttt{LBFGSAttack} \citep{szegedy2013intriguing},
\item \texttt{MomentumIterativeAttack} \citep{dong2018boosting},
\item \texttt{FastFeatureAttack} \citep{sabour2015adversarial}, and
\item \texttt{SpatialTransformAttack} \citep{xiao2018spatially}.
\end{itemize}

Each of the attacks contains three core components:
\begin{itemize}
\item a \predict function,
\item a loss function \loss, and
\item a \perturb method.
\end{itemize}

Taking untargeted \texttt{LinfPGDAttack} on classifiers as the running example, \predict is the classifier, \loss is the loss function for gradient calculation,
the \perturb method takes \x and \y as its arguments, where \x is the input to be attacked, \y is the true label of \x. \texttt{predict(x)} contains the ``logits'' of the neural work. The \loss could be the cross-entropy loss function or another suitable loss function who takes \texttt{predict(x)} and \y as its arguments.

However, the decoupling of these three core components is flexible enough to allow more versatile attacks. In general, we require the \predict and \loss to be designed such that \loss always takes \texttt{predict(x)} and \y as its inputs. As such, no knowledge about \predict and \loss is required by the \perturb method. For example, \texttt{FastFeatureAttack} and \texttt{LinfPGDAttack} share the same underlying \texttt{perturb\_iterative} function, but differ in the \predict and \loss. In \texttt{FastFeatureAttack}, the \texttt{predict(x)} outputs the feature representation from a specific layer, the \y is the \texttt{guide} feature representation that we want \texttt{predict(x)} to match, and the \loss becomes the mean squared error.

More generally, \y could be any targets of the adversarial perturbation, \texttt{predict(x)} can output more complex data structures, as long as the \loss can take them as its inputs. For example, we might want to generate one perturbation that fools both model A's classification result and model B's feature representation at the same time. In this case, we just need to make \y and \texttt{predict(x)} to be tuples of labels and features, and modify the \loss accordingly. There is no need to modify the original perturbation implementation.

\subsection{Other Attacks}
\label{sec:other_attack}

Besides gradient-based attacks, the current version of \advertorch also implements
\begin{itemize}
\item \texttt{SinglePixelAttack}, \texttt{LocalSearchAttack} \citep{narodytska2016simple}, and
\item \texttt{JacobianSaliencyMapAttack} \citep{papernot2016limitations}.
\end{itemize}

\subsection{BPDA Wrapper}
\label{sec:bpda}

The Backward Pass Differentiable Approximation \citep{athalye2018obfuscated} is an attack technique that enhances gradient-based attacks, when attacking defended models who have non-differentiable or gradient-obfuscating components. Specifically, let $y = f(x)$ be a classifier, and let $\hat x = d(x)$ be a preprocessing based defense module that takes the original input $x$ and preprocesses it to be $\hat x$. When $d(\cdot)$ is non-differentiable or gradient-obfuscating, gradient-based attacks will be ineffective on the defended model $f(d(\cdot))$, since $\nabla_x (f(d(x)))$ is either unavailable or uninformative. BPDA solves this problem by replacing the backward pass of $d(\cdot)$, $\frac{\partial d}{\partial x}$, with the backward pass of another function $g(\cdot)$,  $\frac{\partial g}{\partial x}$.

In \advertorch, we implement \texttt{BPDAWrapper} that allows convenient backward pass replacements. With \texttt{BPDAWrapper}, one can either
\begin{itemize}
\item specify $g(\cdot)$, the forward pass function that is used create the backward pass replacement, $\frac{\partial g}{\partial x}$, or
\item directly specify the backward pass replacement $\frac{\partial g}{\partial x}$.
\end{itemize}
To give a concrete example, let \texttt{defense} be the defense module $d(x)$ that preprocesses the input.
\texttt{defense\_withbpda = BPDAWrapper(defense, forwardsub=lambda x: x)} directly returns a defense module with the same forward pass, but having its backward pass replaced with the backward pass of \texttt{forwardsub}, which is specified as the identity function by the function \texttt{lambda x: x}. This specific backward pass replacement is also known as the straight-through gradient estimator \citep{bengio2013estimating}.

\section{Defenses and Robust Training}
\label{sec:defense}

\paragraph{Preprocessing-based Defenses:} The current version implements a few preprocessing based defenses
\footnote{We will keep expanding this list over time.}
 including
\begin{itemize}
\item \texttt{JPEGFilter} \citep{dziugaite2016study},
\item \texttt{BitSqueezing}, \texttt{MedianSmoothing2D} \citep{xu2017feature0}, and
\item linear filters including \texttt{ConvSmoothing2D}, \texttt{AverageSmoothing2D}, and \texttt{GaussianSmoothing2D}.
\end{itemize}

These defenses are all implemented as PyTorch modules which can be easily combined on the fly, thanks to the dynamic computation graph nature of PyTorch.

\paragraph{Robust Training:} Adversarially augmented training \citep{kurakin2016adversarial,madry2017towards,ding2018max} and provably robust training \citep{wong2018provable,mirman2018differentiable,gowal2018effectiveness} have been shown to be the most effective methods against worst-case perturbations. Currently, these training algorithms are not standardized yet,
and usually have different variations that are difficult to be modularized. Therefore, our plan is to provide reference implementations of representative training algorithms in the folder \texttt{advertorch\_examples}. The current version includes an example, \texttt{tutorial\_train\_mnist.py}, implementing \citeauthor{madry2017towards} adversarial training on the MNIST dataset.

\section{Versioning and Reporting Benchmark Results}
\label{sec:version}

\advertorch follows Semantic Versioning 2.0.0 \citep{preston2013semantic}, where the version number takes the \texttt{MAJOR.MINOR.PATCH} format. Given such a version number, quoting from \citet{preston2013semantic}, we increment the:
\begin{enumerate}
\item \texttt{MAJOR} version when we make incompatible API changes,
\item \texttt{MINOR} version when we add functionality in a backwards-compatible manner, and
\item \texttt{PATCH} version when we make backwards-compatible bug fixes.
\end{enumerate}

When benchmark reporting results from \advertorch, the authors should report the MAJOR.MINOR version number and detailed hyperparameters. For example, when performing untargeted \texttt{LinfPGDAttack}, the following hyperparameters shall be reported: the loss function, the maximum perturbation magnitude, the number of iterations, the step size, and whether the attack is randomly initialized.

\section*{Acknowledgments}

We thank April Cooper for designing the logo for \advertorch and Marcus Brubaker for helpful discussions.

\bibliographystyle{apalike}
\bibliography{refs}

\begin{thebibliography}{}

\bibitem[Athalye et~al., 2018]{athalye2018obfuscated}
Athalye, A., Carlini, N., and Wagner, D. (2018).
\newblock Obfuscated gradients give a false sense of security: Circumventing
  defenses to adversarial examples.
\newblock In {\em International Conference on Machine Learning}, pages
  274--283.

\bibitem[Bengio et~al., 2013]{bengio2013estimating}
Bengio, Y., L{\'e}onard, N., and Courville, A. (2013).
\newblock Estimating or propagating gradients through stochastic neurons for
  conditional computation.
\newblock {\em arXiv preprint arXiv:1308.3432}.

\bibitem[Biggio et~al., 2013]{biggio2013data}
Biggio, B., Pillai, I., Rota~Bul{\`o}, S., Ariu, D., Pelillo, M., and Roli, F.
  (2013).
\newblock Is data clustering in adversarial settings secure?
\newblock In {\em Proceedings of the 2013 ACM workshop on Artificial
  intelligence and security}, pages 87--98. ACM.

\bibitem[Brendel et~al., 2017]{brendel2017decision}
Brendel, W., Rauber, J., and Bethge, M. (2017).
\newblock Decision-based adversarial attacks: Reliable attacks against
  black-box machine learning models.
\newblock {\em arXiv preprint arXiv:1712.04248}.

\bibitem[Carlini and Wagner, 2017]{carlini2017towards}
Carlini, N. and Wagner, D. (2017).
\newblock Towards evaluating the robustness of neural networks.
\newblock In {\em Security and Privacy (SP), 2017 IEEE Symposium on}, pages
  39--57. IEEE.

\bibitem[Cisse et~al., 2017]{cisse2017parseval}
Cisse, M., Bojanowski, P., Grave, E., Dauphin, Y., and Usunier, N. (2017).
\newblock Parseval networks: Improving robustness to adversarial examples.
\newblock In {\em International Conference on Machine Learning}, pages
  854--863.

\bibitem[Ding et~al., 2019]{ding2018on}
Ding, G.~W., Lui, K. Y.-C., Jin, X., Wang, L., and Huang, R. (2019).
\newblock On the sensitivity of adversarial robustness to input data
  distributions.
\newblock In {\em International Conference on Learning Representations}.

\bibitem[Ding et~al., 2018]{ding2018max}
Ding, G.~W., Sharma, Y., Lui, K. Y.~C., and Huang, R. (2018).
\newblock Max-margin adversarial ({MMA}) training: Direct input space margin
  maximization through adversarial training.
\newblock {\em arXiv preprint arXiv:1812.02637}.

\bibitem[Dong et~al., 2018]{dong2018boosting}
Dong, Y., Liao, F., Pang, T., Su, H., Zhu, J., Hu, X., and Li, J. (2018).
\newblock Boosting adversarial attacks with momentum.
\newblock In {\em Proceedings of the IEEE Conference on Computer Vision and
  Pattern Recognition}, pages 9185--9193.

\bibitem[Dziugaite et~al., 2016]{dziugaite2016study}
Dziugaite, G.~K., Ghahramani, Z., and Roy, D.~M. (2016).
\newblock A study of the effect of jpg compression on adversarial images.
\newblock {\em arXiv preprint arXiv:1608.00853}.

\bibitem[Fawzi et~al., 2017]{fawzi2017classification}
Fawzi, A., Moosavi-Dezfooli, S.-M., Frossard, P., and Soatto, S. (2017).
\newblock Classification regions of deep neural networks.
\newblock {\em arXiv preprint arXiv:1705.09552}.

\bibitem[Feinman et~al., 2017]{feinman2017detecting}
Feinman, R., Curtin, R.~R., Shintre, S., and Gardner, A.~B. (2017).
\newblock Detecting adversarial samples from artifacts.
\newblock {\em arXiv preprint arXiv:1703.00410}.

\bibitem[Goodfellow et~al., 2014]{goodfellow2014explaining}
Goodfellow, I.~J., Shlens, J., and Szegedy, C. (2014).
\newblock Explaining and harnessing adversarial examples.
\newblock {\em arXiv preprint arXiv:1412.6572}.

\bibitem[Gowal et~al., 2018]{gowal2018effectiveness}
Gowal, S., Dvijotham, K., Stanforth, R., Bunel, R., Qin, C., Uesato, J., Mann,
  T., and Kohli, P. (2018).
\newblock On the effectiveness of interval bound propagation for training
  verifiably robust models.
\newblock {\em arXiv preprint arXiv:1810.12715}.

\bibitem[Guo et~al., 2017]{guo2017countering}
Guo, C., Rana, M., Cisse, M., and van~der Maaten, L. (2017).
\newblock Countering adversarial images using input transformations.
\newblock {\em arXiv preprint arXiv:1711.00117}.

\bibitem[Katz et~al., 2017]{katz2017reluplex}
Katz, G., Barrett, C., Dill, D.~L., Julian, K., and Kochenderfer, M.~J. (2017).
\newblock Reluplex: An efficient smt solver for verifying deep neural networks.
\newblock In {\em International Conference on Computer Aided Verification},
  pages 97--117. Springer.

\bibitem[Kurakin et~al., 2016]{kurakin2016adversarial}
Kurakin, A., Goodfellow, I., and Bengio, S. (2016).
\newblock Adversarial machine learning at scale.
\newblock {\em arXiv preprint arXiv:1611.01236}.

\bibitem[Ma et~al., 2018]{ma2018characterizing}
Ma, X., Li, B., Wang, Y., Erfani, S.~M., Wijewickrema, S., Houle, M.~E.,
  Schoenebeck, G., Song, D., and Bailey, J. (2018).
\newblock Characterizing adversarial subspaces using local intrinsic
  dimensionality.
\newblock {\em arXiv preprint arXiv:1801.02613}.

\bibitem[Madry et~al., 2017]{madry2017towards}
Madry, A., Makelov, A., Schmidt, L., Tsipras, D., and Vladu, A. (2017).
\newblock Towards deep learning models resistant to adversarial attacks.
\newblock {\em arXiv preprint arXiv:1706.06083}.

\bibitem[Metzen et~al., 2017]{metzen2017detecting}
Metzen, J.~H., Genewein, T., Fischer, V., and Bischoff, B. (2017).
\newblock On detecting adversarial perturbations.
\newblock {\em arXiv preprint arXiv:1702.04267}.

\bibitem[Mirman et~al., 2018]{mirman2018differentiable}
Mirman, M., Gehr, T., and Vechev, M. (2018).
\newblock Differentiable abstract interpretation for provably robust neural
  networks.
\newblock In Dy, J. and Krause, A., editors, {\em Proceedings of the 35th
  International Conference on Machine Learning}, volume~80 of {\em Proceedings
  of Machine Learning Research}, pages 3575--3583, Stockholmsmässan, Stockholm
  Sweden. PMLR.

\bibitem[Narodytska and Kasiviswanathan, 2016]{narodytska2016simple}
Narodytska, N. and Kasiviswanathan, S.~P. (2016).
\newblock Simple black-box adversarial perturbations for deep networks.
\newblock {\em arXiv preprint arXiv:1612.06299}.

\bibitem[Papernot et~al., 2016a]{papernot2016transferability}
Papernot, N., McDaniel, P., and Goodfellow, I. (2016a).
\newblock Transferability in machine learning: from phenomena to black-box
  attacks using adversarial samples.
\newblock {\em arXiv preprint arXiv:1605.07277}.

\bibitem[Papernot et~al., 2016b]{papernot2016limitations}
Papernot, N., McDaniel, P., Jha, S., Fredrikson, M., Celik, Z.~B., and Swami,
  A. (2016b).
\newblock The limitations of deep learning in adversarial settings.
\newblock In {\em Security and Privacy (EuroS\&P), 2016 IEEE European Symposium
  on}, pages 372--387. IEEE.

\bibitem[Paszke et~al., 2017]{paszke2017automatic}
Paszke, A., Gross, S., Chintala, S., Chanan, G., Yang, E., DeVito, Z., Lin, Z.,
  Desmaison, A., Antiga, L., and Lerer, A. (2017).
\newblock Automatic differentiation in pytorch.
\newblock In {\em NIPS-W}.

\bibitem[Preston-Werner, 2013]{preston2013semantic}
Preston-Werner, T. (2013).
\newblock Semantic versioning 2.0.0.
\newblock \url{https://semver.org/\#semantic-versioning-200}.

\bibitem[Rauber et~al., 2017]{rauber2017foolbox}
Rauber, J., Brendel, W., and Bethge, M. (2017).
\newblock Foolbox: A python toolbox to benchmark the robustness of machine
  learning models.
\newblock {\em arXiv preprint arXiv:1707.04131}.

\bibitem[Sabour et~al., 2015]{sabour2015adversarial}
Sabour, S., Cao, Y., Faghri, F., and Fleet, D.~J. (2015).
\newblock Adversarial manipulation of deep representations.
\newblock {\em arXiv preprint arXiv:1511.05122}.

\bibitem[Shafahi et~al., 2019]{shafahi2018adversarial}
Shafahi, A., Huang, W.~R., Studer, C., Feizi, S., and Goldstein, T. (2019).
\newblock Are adversarial examples inevitable?
\newblock In {\em International Conference on Learning Representations}.

\bibitem[Szegedy et~al., 2013]{szegedy2013intriguing}
Szegedy, C., Zaremba, W., Sutskever, I., Bruna, J., Erhan, D., Goodfellow, I.,
  and Fergus, R. (2013).
\newblock Intriguing properties of neural networks.
\newblock {\em arXiv preprint arXiv:1312.6199}.

\bibitem[Weng et~al., 2018]{weng2018evaluating}
Weng, T.-W., Zhang, H., Chen, P.-Y., Yi, J., Su, D., Gao, Y., Hsieh, C.-J., and
  Daniel, L. (2018).
\newblock Evaluating the robustness of neural networks: An extreme value theory
  approach.
\newblock {\em arXiv preprint arXiv:1801.10578}.

\bibitem[Wong and Kolter, 2018]{wong2018provable}
Wong, E. and Kolter, Z. (2018).
\newblock Provable defenses against adversarial examples via the convex outer
  adversarial polytope.
\newblock In {\em International Conference on Machine Learning}, pages
  5283--5292.

\bibitem[Xiao et~al., 2018]{xiao2018spatially}
Xiao, C., Zhu, J.-Y., Li, B., He, W., Liu, M., and Song, D. (2018).
\newblock Spatially transformed adversarial examples.
\newblock {\em arXiv preprint arXiv:1801.02612}.

\bibitem[Xu et~al., 2017]{xu2017feature0}
Xu, W., Evans, D., and Qi, Y. (2017).
\newblock Feature squeezing: Detecting adversarial examples in deep neural
  networks.
\newblock {\em arXiv preprint arXiv:1704.01155}.

\end{thebibliography}

\end{document}